\documentclass[letterpaper, 10 pt, conference]{ieeeconf}  

\IEEEoverridecommandlockouts                              

\usepackage{cite}
\usepackage{amsmath}
\usepackage{cleveref}
\usepackage{amssymb}  

\usepackage{booktabs,multirow}
\usepackage[table]{xcolor}
\usepackage{adjustbox} 

\usepackage{amsfonts}
\usepackage{multirow}
\usepackage{xcolor}
\usepackage{colortbl}
\definecolor{tabfirst}{rgb}{1, 0.7, 0.7}
\definecolor{tabsecond}{rgb}{1, 0.85, 0.7}
\definecolor{tabthird}{rgb}{1, 1, 0.7}
\newcommand{\etc}{\text{etc.}}
\newcommand{\eg}{\text{e.g.}}

\title{\LARGE \bf
ArmGS: Composite Gaussian Appearance Refinement \\ for Modeling Dynamic Urban Environments
}

\author{Guile Wu$^{1}$, Dongfeng Bai$^{1}$, and Bingbing Liu$^{1}$
\thanks{$^{1}$The authors are with Huawei Noah's Ark Lab.
{\{guile.wu, baidongfeng, liu.bingbing\}@huawei.com}}%
}

\begin{document}

\maketitle
\thispagestyle{empty}
\pagestyle{empty}

\begin{abstract}
This work focuses on modeling dynamic urban environments for autonomous driving simulation.
Contemporary data-driven methods using neural radiance fields have achieved photorealistic driving scene modeling, but they suffer from low rendering efficacy.
Recently, some approaches have explored 3D Gaussian splatting for modeling dynamic urban scenes, enabling high-fidelity reconstruction and real-time rendering.
However, these approaches often neglect to model fine-grained variations between frames and camera viewpoints, leading to suboptimal results.
In this work, we propose a new approach named ArmGS that exploits composite driving Gaussian splatting with multi-granularity appearance refinement for autonomous driving scene modeling.
The core idea of our approach is devising a multi-level appearance modeling scheme to optimize a set of transformation parameters for composite Gaussian refinement from multiple granularities, ranging from local Gaussian level to global image level and dynamic actor level.
This not only models global scene appearance variations between frames and camera viewpoints, but also models local fine-grained changes of background and objects.
Extensive experiments on multiple challenging autonomous driving datasets, namely, Waymo, KITTI, NOTR and VKITTI2, demonstrate the superiority of our approach over the state-of-the-art methods.
\end{abstract}


\section{INTRODUCTION}
\label{sec:intro}
Autonomous driving has made remarkable advancements in recent years, but how to effectively validate the safety and reliability of an autonomous driving system remains an open question.
In real-world scenarios, there are many corner cases which cannot be easily collected but are critical to the safety of autonomous driving.
Simulation is one of the most useful ways to alleviate this problem.
It enables the reconstruction and generation of various challenging driving scenes for downstream closed-loop evaluation.

\begin{figure}
    \centering
    \includegraphics[width=0.99\columnwidth]{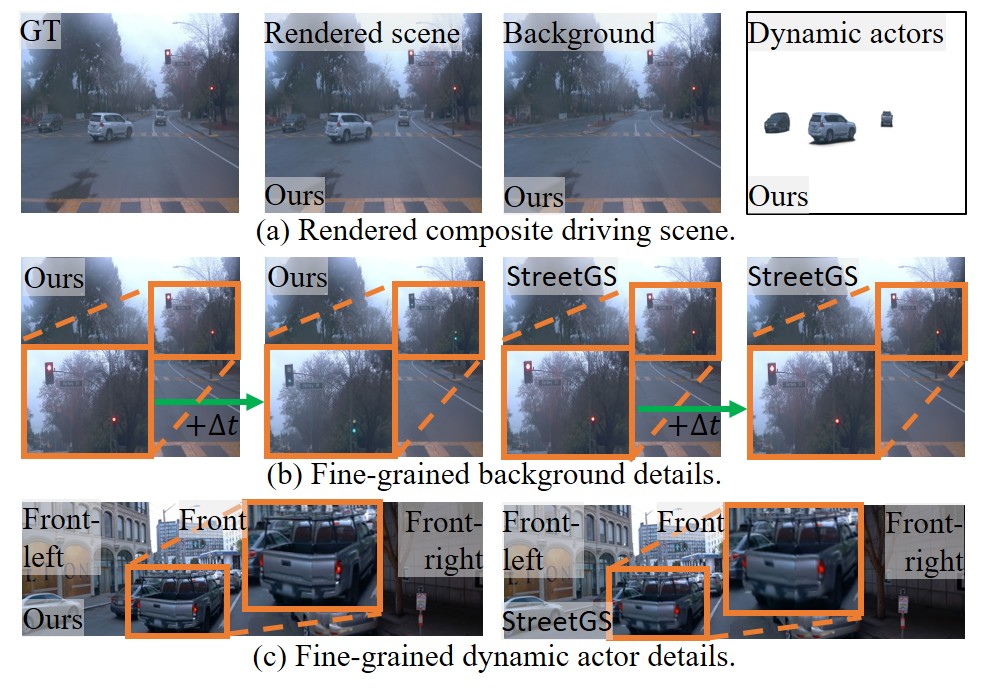}
    \caption{
        An illustration of our ArmGS for modeling urban scenes.
        Our approach is capable of modeling fine-grained changes of background scenes and objects across frames and camera viewpoints.
        }
    \label{fig:illustration}
\end{figure}

Traditionally, virtual-engine-based methods, such as CARLA~\cite{dosovitskiy2017carla}, use classic graphics rendering systems for driving scene simulation, but the simulated scenes often have significant gaps from the real-world ones.
On the other hand, data-driven simulation enables photorealistic driving scene modeling in a low-cost and efficient way~\cite{yan2024street,yang2024emernerf,chen2025omnire,huang2024textit,yang2023unisim,wu2023mars}.
With the great success of Neural Radiance Fields (NeRFs) for neural rendering,
some researchers have adapted NeRFs for scene modeling in autonomous driving~\cite{yang2024emernerf,yang2023unisim,wu2023mars,ost2021neural}.
Although NeRF-based methods have shown great potential for reconstructing realistic driving scenes, they mostly suffer from long training time and fail to achieve real-time rendering. 
More recently, some researchers have resorted to 3D Gaussian Splatting (3DGS)~\cite{kerbl20233d} to learn explicit 3D scene representations for driving scene modeling~\cite{yan2024street,huang2024textit,chen2025omnire,miao2025evolsplat,chen2023periodic,zhou2024drivinggaussian}.
Due to the efficiency of explicit representations and differentiable tile rasterizers, 3DGS-based methods have shown promising results for high-fidelity reconstruction and real-time rendering.
However, existing approaches for driving scene modeling mostly neglect to model fine-grained variations across frames and camera viewpoints.
In autonomous driving simulation, scene images are captured in the wild along the movement of the ego vehicle.
This unavoidably brings significant appearance changes of scenes and objects across frames and camera viewpoints due to lighting and camera exposure.
Thus, failing to model these variations can lead to the loss of fine-grained details and suboptimal results.

In this work, we propose to exploit composite 3D \textbf{G}aussian \textbf{S}platting with \textbf{A}ppearance \textbf{R}efine\textbf{M}ent (ArmGS) from multiple granularities for dynamic urban environments modeling in autonomous driving.
Our approach constructs a composite driving scene model based on 3DGS representations and employs a multi-level appearance modeling scheme to optimize a set of transformation parameters for composite Gaussian refinement.
Specifically, we represent a driving scene with composite 3DGS and refine 3DGS at local Gaussian level, global image level and dynamic actor level, respectively.
For local Gaussian level refinement, we extract latent Gaussian appearance representation to learn Gaussian-wise transformation parameters for Gaussian appearance refinement.
This helps to learn local fine-grained appearance variations in the 3D Gaussian space.
For global image level refinement, we extract latent image appearance representation to learn global image-wise transformation parameters for image color refinement. 
This helps to learn global image-wise appearance changes across frames and camera viewpoints.
Moreover, to model dynamic variations of moving road actors (such as vehicles), we employ a light-weight spatial-temporal deformation model to transform dynamic actor Gaussians across time.
This multi-level appearance modeling scheme allows for refining the composite driving scene Gaussians at multiple granularities.
Consequently, scene and road actors appearance variations are collectively modeled to reconstruct richer fine-grained cues of driving scenes.
Our experiments on multiple challenging autonomous driving datasets, namely, Waymo~\cite{sun2020scalability}, KITTI~\cite{geiger2012we}, NOTR~\cite{yang2024emernerf} and VKITTI2~\cite{cabon2020virtual}, for driving scene modeling and novel view synthesis show that our approach not only reconstructs global scene appearance variations between frames and camera viewpoints, but also models local fine-grained changes of background and dynamic objects.
An illustration is shown in Fig.~\ref{fig:illustration}.
In summary, our \textbf{contributions} are:
\textbf{(I)} {Our method is the first method that proposes to explicitly embed multi-granularity appearances to model fine-grained changes for dynamic urban environments modeling, which fills a gap left by existing methods;}
\textbf{(II)} {Our method employs a set of transformation parameters for composite Gaussian refinement, which is simple, effective, and preserves the differentiability of the splatting process;}
\textbf{(III)} {Our thorough experiments on Waymo, KITTI, NOTR and VKITTI2 show the superiority of our method over the state-of-the-art methods.}

\begin{figure*}[t]
    \centering
    \includegraphics[width=0.85\textwidth]{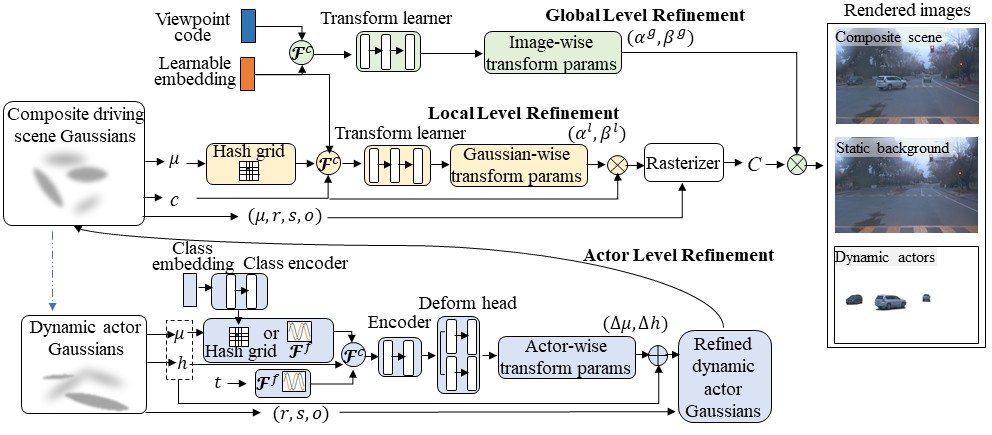}
    \caption{An overview of our approach.
    Our approach refines composite driving scene Gaussians with appearance modeling at multiple granularities, ranging from local Gaussians level to global images level and dynamic actors level.
    The modules for local level refinement, global level refinement and actor level refinement are indicated in yellow, green and blue, respectively.
    }
    \label{fig:framework}
\end{figure*}

\section{RELATED WORK}
\label{sec:related-work}

\subsection{3D Gaussian Splatting}
3D Gaussian Splatting (3DGS)~\cite{kerbl20233d} is a recently proposed real-time rendering approach.
Unlike NeRFs that employ a continuous radiance field to implicitly represent 3D scenes, 3DGS represents a 3D scene with a collection of explicit 3D Gaussian primitives and utilizes splatting-based rasterization~\cite{zwicker2001ewa} to project 3D Gaussians to 2D image space.
Since the success of~\cite{kerbl20233d}, researchers have made many efforts to improve 3DGS from different aspects~\cite{feng2025flashgs,yu2024mip,zhang2024gaussian,dahmani2024swag,qian2025weathergs}, such as anti-aliasing~\cite{yu2024mip}, compression~\cite{fan2024lightgaussian}, surface reconstruction~\cite{lyu20243dgsr}, etc.
Among them, SWAG~\cite{dahmani2024swag} proposes to model image appearance changes with image embeddings for novel-view synthesis from unconstrained image collections.
Our approach significantly differs from~\cite{dahmani2024swag} in that we devise a novel multi-level appearance modeling scheme to optimize a set of transformation parameters for composite Gaussian refinement, rather than modeling transient objects and landmarks from unconstrained images.
In addition, 4DGS~\cite{wu20244d} proposes to explore deformation fields with HexPlane representations for real-time dynamic 4D scene rendering, which is subsequently modified by~\cite{huang2024textit} for dynamic urban scene reconstruction.
Although we also employ deformation fields for dynamic actor level refinement, our method is different in that we construct a light-weight network structure and encode position, timestamp and spherical harmonics to learn actor variations, rather than constructing HexPlane representation~\cite{wu20244d,huang2024textit}.

\subsection{Autonomous Driving Scene Modeling}
Autonomous driving simulation~\cite{dosovitskiy2017carla,yang2023unisim,yan2024street,yang2024emernerf,liu2024vqa,liu2023mv} is critical for real-world closed-loop evaluation.
For dynamic driving scene modeling, there are primarily two types of methods, namely, virtual-engine-based methods~\cite{dosovitskiy2017carla,cabon2020virtual} and data-driven methods~\cite{yan2024street,yang2024emernerf,yang2023unisim,miao2025evolsplat}.
While virtual-engine-based methods typically lack scalability and suffer from significant domain gaps, data-driven methods are becoming prevailing in recent years due to their photo-realism, scalability and efficiency.
For data-driven simulation, NeRF-based methods~\cite{fischer2024multi,yang2024emernerf,ost2021neural,guo2023streetsurf} have shown compelling performance, but require point dense sampling along each ray to model the interaction between light and scene elements.
This leads to expensive computational costs and low rendering speed.
On the other hand, 3DGS-based methods~\cite{yan2024street,huang2024textit,chen2025omnire,fischer2024dynamic,ren2024unigaussian} make full use of the advantages of 3DGS and achieve real-time rendering for modeling driving scenes.
StreetGS~\cite{yan2024street} proposes to model dynamic urban scenes with road asset pose optimization and 4D spherical harmonics appearance model, but neglects to model fine-grained driving scene variations between frames and camera viewpoints.
S$^3$Gaussian\cite{huang2024textit} extends dynamic 4DGS~\cite{wu20244d} for driving scene reconstruction in a self-supervised manner, but the rendering results are suboptimal with some artifacts and do not explicitly model fine-grained changes of driving scenes.
Our work also builds on 3DGS, but different from existing works, our approach refines composite driving scene Gaussians with appearance modeling at multiple granularities, ranging from local Gaussians to global images and dynamic actors.
Our approach is capable of modeling global scene appearance variations, local fine-grained changes of background and dynamic objects in autonomous driving scenes.

\section{METHODOLOGY}
\label{sec:method}

\subsection{Preliminaries}
\paragraph{3D Gaussian Splatting}
Formally, 3DGS~\cite{kerbl20233d} consists of a collection of learnable parameters, where each Gaussian is defined by the position (mean) parameter $\mu\in\mathbb{R}^3$, the covariance matrix $\Sigma$ (which is determined by the rotation parameter $r\in\mathbb{R}^4$ and the scaling parameter $s\in\mathbb{R}^3$), the opacity parameter $o\in\mathbb{R}^1$ and the spherical harmonics parameter $h\in\mathbb{R}^d$ ($d$ is determined by the degrees of spherical harmonics).
These parameters collectively represent Gaussians in a 3D space as:
\begin{equation}
\label{eq:gaussian}
G(x) = e^{-\frac{1}{2}(x-\mu)^{T}\Sigma^{-1}(x-\mu)},
\end{equation}
where point $x$ is centered at $\mu$.
To render images from different camera viewpoints, 3D Gaussians are projected onto the corresponding image plane and the color $C$ of each pixel is calculated with $N$-ordered 2D splats using $\alpha$-blending as:
\begin{equation}
\label{eq:gaussian-bleding}
C = \sum_{i{\in}N}c_i\alpha_i\prod_{j=1}^{i-1}(1-\alpha_j),
\end{equation}
where $c_i$ is the color derived from $h$ and $\alpha_i$ is determined by $o$ and the contribution of the Gaussian.

\paragraph{Composite Driving Scene Gaussians}
Inspired by~\cite{ost2021neural,yan2024street,zhou2024hugs}, we represent a driving scene with composite 3DGS models, including a background Gaussian model, dynamic actor Gaussian models and a sky Gaussian model.
In the background Gaussian model, we use a set of 3D Gaussians to model the background scene in the world coordinate space.
For each dynamic actor (such as moving vehicle) in a driving scene, we represent it with a separate Gaussian model in the object-centric coordinate space.
These dynamic actor Gaussians are transformed to world coordinates with pose transformation matrices to form the composite scene Gaussians.
For distant sky region, we construct the sky Gaussian model with explicit cubemap representations~\cite{yan2024street} and model sky color from view direction.
These models collectively form the composite Gaussians to be rendered via a differentiable tile-based rasterizer.

Although composite scene Gaussians can reconstruct 3D driving scenes in an efficient way, existing methods~\cite{yan2024street,ren2024unigaussian,zhou2024hugs} lack the ability to effectively model driving scene appearance variations between frames and camera viewpoints, leading to missing fine-grained details and suboptimal simulation results. 
To resolve this problem, we present composite 3DGS appearance refinement via a multi-level appearance modeling scheme.

\subsection{The Proposed Approach}
An overview of our ArmGS approach is depicted in Fig.~\ref{fig:framework}.
Our approach realizes driving scene appearance refinement from three levels, namely, local composite Gaussian level, global image level and dynamic actor level.

\paragraph{Local Composite Gaussian Level Refinement}
With composite driving Gaussians, we propose to model fine-grained scene variations by learning a set of transformation parameters to refine composite Gaussians.
First, to model changes across scene frames, we construct a low-dimensional learnable embedding $\epsilon$ for each frame.
Then, we employ a multiresolution Hash grid $\mathcal{H}$~\cite{muller2022instant} to extract richer Gaussian position code and merge the low-dimensional embedding, the position code and the Gaussian color as the \emph{latent Gaussian appearance representation} $f^l$ of each Gaussian.
This is formulated as:
\begin{equation}
\label{eq:latent_gaussian_feat}
f^l = \mathcal{F}^c(\mathcal{H}(\mu), \epsilon, c),
\end{equation}
where the merge function $\mathcal{F}^c$ is concatenation in this work.
Then, we employ a local Gaussian appearance affine transformation learner $\mathcal{D}^l$ (a light-weight MLP) with $f^l$ as the input to learn local Gaussian-wise transformation parameters $(\alpha^l\in\mathbb{R}^3, \beta^l\in\mathbb{R}^3)$ for Gaussian appearance refinement, which is defined as:
\begin{equation}
\label{eq:transformation_gaussian}
c' = \alpha^{l}c+\beta^{l},
\end{equation}
where $(\alpha^l, \beta^l) = \mathcal{D}^l(f^l)$.
$c'$ is then used in Eq.\eqref{eq:gaussian-bleding} for color computation.
In this way, the contribution of each Gaussian to the rendered image is modulated across frames, so fine-grained scene variations are effectively encoded into local Gaussian-wise appearance variations across frames.
During rendering, to obtain a low-dimensional embedding $\epsilon$ for a novel view, we consider that the appearance variation of a novel view should be consistent with its adjacent frame.
Thus, for each novel view, we calculate the index of its nearest neighbor training frames based on camera index and timestamp to get its corresponding $\epsilon$, and then use Eqs.\eqref{eq:latent_gaussian_feat} and \eqref{eq:transformation_gaussian} for the transformation.

\paragraph{Global Image Level Refinement}
Although the local Gaussian-wise appearance refinement encourages each Gaussian to learn local fine-grained variations, it does not encourage global consistent variations across camera viewpoints.
For example, image-wise appearance variation can be consistently changed due to direct sunlight or camera exposure,
which cannot be resolved by image preprocessing and requires the holistic scene appearance refinement.
Thus, to compensate for the global image-wise variation, we further refine driving scene appearance at the global image level.
Specifically, we encode camera location and viewing direction as a camera viewpoint code $\phi$
and merge it with the embedding $\epsilon$ as the \emph{latent image appearance representation} $f^g$, which is defined as:
\begin{equation}
\label{eq:latent_image_feat}
f^g = \mathcal{F}^c(\epsilon, \phi).
\end{equation}
We then employ an image appearance affine transformation learner $\mathcal{D}^g$ with $f^g$ as the input to learn global image-wise transformation parameters $(\alpha^g\in\mathbb{R}^{3{\times}3}, \beta^g\in\mathbb{R}^{1{\times}3)}$ for image color transformation, which is formulated as:
\begin{equation}
\label{eq:transformation_image}
C' = \alpha^{g}C+\beta^{g},
\end{equation}
where $(\alpha^g, \beta^g) = \mathcal{D}^g(f^g)$.
With the collaboration of local Gaussian-wise appearance modeling and global image-wise appearance modeling, the composite driving scene Gaussians are modulated at different granularities.

\paragraph{Dynamic Actor Level Refinement}
In dynamic driving scene modeling, dynamic road actors usually have more complex motion and appearance variations than the background and static objects.
For example, a moving vehicle needs to be placed at different locations along its trajectory, while its appearance is affected not only by the scene but also by its actions, such as using brake signals.
To deal with this problem, we also refine composite driving Gaussians at the dynamic actor level.
Specifically, we employ a low-dimensional learnable embedding for each class of dynamic actors and construct a light-weight class encoder to learn the weights of the dynamic actor Hash grid, inspired by~\cite{ha2016hypernetworks,yang2023unisim}.
This helps to encode class information into the Hash grid for position encoding.
Alternatively, to reduce training parameters, we observe that the simple yet effective sinusoidal encoding~\cite{mildenhall2021nerf} also works well.
Then, to model the complex appearance variations specific to each dynamic actor, we construct a light-weight spatial-temporal encoder and a light-weight deformation head to transform position and spherical harmonics attributes of dynamic actor Gaussians across time.
Instead of using the HexPlane~\cite{wu20244d} to learn the spatial-temporal structure, we encode position $\mu$ with the encoding function $\mathcal{F}^a$ (class-wise Hash grid or sinusoidal encoding), timestamp $t$ with a sinusoidal function $\mathcal{F}^f$ and merge them with the spherical harmonics $h$ to learn the \emph{spatial-temporal actor representation} $f^a$ of the dynamic actor.
This is defined as:
\begin{equation}
\label{eq:spatial_temporal}
f^a=\mathcal{D}^a(\mathcal{F}^c(\mathcal{F}^a(\mu), \mathcal{F}^f(t), h)),
\end{equation}
where $\mathcal{D}^a$ is a shared spatial-temporal representation encoder.
Then, we use a deformation head (a multi-head MLP) $\mathcal{D}^h$ with $f^a$ as the input to learn deformation for Gaussian position and color refinement, as:
\begin{equation}
\label{eq:deformation}
\{\mu{'}, h{'}\}{=}\{\mu{+}\Delta{\mu}, h{+}\Delta{h}\},
\end{equation}
where $\{\Delta{\mu}, \Delta{h}\}{=}\mathcal{D}^h(f^a)$.
Note that, although Gaussian deformation is inspired by~\cite{wu20244d}, our design differs from existing works~\cite{wu20244d,huang2024textit,fischer2024dynamic} in that we construct light-weight encoders and deformation heads and encodes position, timestamp and spherical harmonics to learn actor variations, rather than constructing HexPlane representation~\cite{wu20244d,huang2024textit} or using neural fields for color computation~\cite{fischer2024dynamic}.
Moreover, the collaboration of the local level refinement, the global level refinement and the dynamic actor level refinement brings superior performance for dynamic urban environments modeling in autonomous driving.

\subsection{Composite Driving Gaussian Optimization}
\paragraph{Training Objective}
We optimize our approach in an end-to-end differentiable rendering manner.
The training objective loss is defined as:
\begin{equation}
\label{eq:loss_overall}
\mathcal{L}=(1-\lambda_{1})\mathcal{L}_{rgb}+\lambda_{1}\mathcal{L}_{ssim}+\lambda_{2}\mathcal{L}_{d}+\lambda_{3}\mathcal{L}_{s}+\lambda_{4}\mathcal{L}_{f},
\end{equation}
where $\lambda_i$ is the weight coefficient, $\mathcal{L}_{rgb}$ and $\mathcal{L}_{ssim}$ are the image reconstruction loss between rendering images and ground-truth images following~\cite{kerbl20233d}, $\mathcal{L}_{d}$ is a depth loss calculated by a L1 loss between rendering depth maps and LiDAR depth, $\mathcal{L}_{s}$ is a sky mask loss computed by a binary cross-entropy loss between rendering sky masks and pre-extracted sky masks~\cite{ren2024grounded}, and $\mathcal{L}_{f}$ is a foreground decomposition loss~\cite{yan2024street} calculated by an entropy loss on the accumulated alpha values of dynamic actors.

\paragraph{Actor Pose Optimization}
For pose transformation parameters (the rotation quaternion $R_t$ and the translation $T_t$) of each actor, we set them as learnable parameters and initialize them with the provided tracked boxes.
The position and rotation of each Gaussian of each actor in the world space are defined as
$\mu=R_t\mu^{a}+T_t$ and $r=R_tr^{a}$, where $\mu^{a}$ and $r^{a}$ are dynamic actor poses in the object-centric coordinate space.
These transformation parameters are directly optimized with scene reconstruction through gradient back-propagation.
When rendering novel views, we interpolate actor poses based on timestamps and the optimized poses.

\begin{table*}[ht!]
    \small
    \centering
    \caption{Quantitative comparison with the state-of-the-art methods on Waymo and KITTI.    
    }
    \begin{tabular}{l|rrr|rrr|rrr}
        \hline
        \multirow{2}{*}{Method} 
        & \multicolumn{3}{c|}{Waymo Reconstruction} & \multicolumn{3}{c|}{Waymo Novel View}& \multicolumn{3}{c}{KITTI}\\
        & PSNR$\uparrow$ & SSIM$\uparrow$ & LPIPS$\downarrow$ & PSNR$\uparrow$ & SSIM$\uparrow$ & LPIPS$\downarrow$ & PSNR$\uparrow$ & SSIM$\uparrow$ & LPIPS$\downarrow$ \\
        \hline
        \hline        
        NSG~\cite{ost2021neural}             & -          & -         & -            & 28.3           & 0.862          & 0.346             & 21.3          & 0.659         & 0.266 \\        
        SUDS~\cite{turki2023suds}            & -          & -         & -             & -          & -         & -            & 23.1          & 0.821         & 0.135 \\
        3DGS~\cite{kerbl20233d}            & 32.5       & 0.938     & 0.089        & 29.6           & 0.918          & 0.117             & 19.2          & 0.739         & 0.174 \\
        S$^3$Gaussian~\cite{huang2024textit}   & 35.6       & 0.946     & 0.088        & 32.8           & 0.931          & 0.097             & -          & -         & - \\
        EmerNeRF~\cite{yang2024emernerf}        & 34.5       & 0.913     & 0.123       & 30.9           & 0.905          & 0.133             & -          & -         & -             \\
        ML-NSG~\cite{fischer2024multi}          & -         & -         & -             & -          & -         & -            & 27.5          & 0.898         & 0.055 \\
        StreetGS~\cite{yan2024street}        & {36.3}      & {0.948}   & {0.074}     & {34.6}    & 0.938     & {0.079}         & 27.8          & 0.889         & 0.069 \\
        OmniRe~\cite{chen2025omnire}          & 35.9       & 0.953     & 0.108        & 31.9      & 0.899     & 0.126         & -      & -            & - \\
        SplatFlow~\cite{sun2025splatflow}       & -         & -         & -             & -          & -         & -            & 27.9         & 0.927         & 0.093 \\        
        \hline
        ArmGS (Ours)    & \textbf{38.1}  &\textbf{0.957} & \textbf{0.064}   & \textbf{35.7}  & \textbf{0.944} & \textbf{0.074}    & \textbf{30.3}          & \textbf{0.931}         & \textbf{0.036} \\
        \hline
    \end{tabular}    
    \label{tab:exp-sota-waymo-kitti}
\end{table*}

\begin{figure*}[t]
    \centering
    \includegraphics[width=0.99\textwidth]{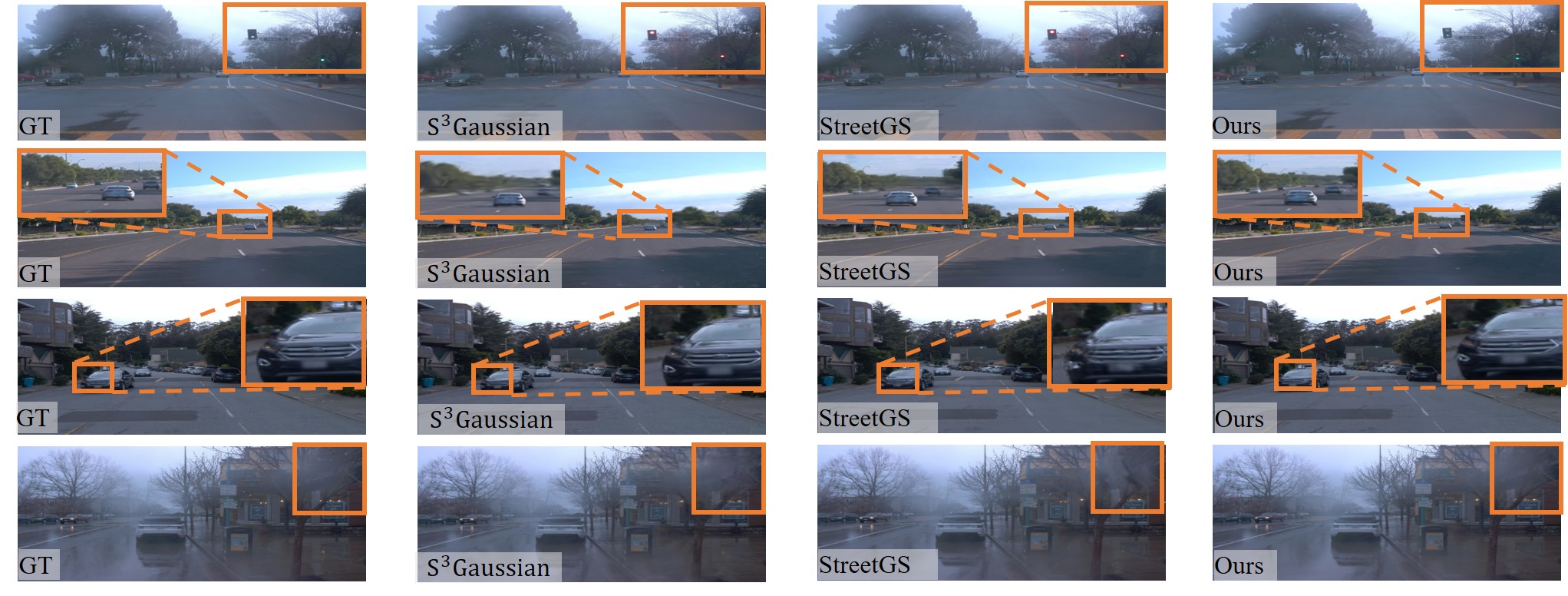}
    \caption{Qualitative comparison with the state-of-the-arts on Waymo.
    From the first to the fourth row, we show results of scene modeling under foggy, sunny, cloudy, and rainy conditions.
    We highlight some fine-grained details, \eg, traffic lights, vehicles and trees.
    }
    \label{fig:exp_vis_waymo}
\end{figure*}

\section{EXPERIMENT}
\label{sec:exp}

\subsection{Datasets and Evaluation Protocol}
\label{sec:exp-dataset}

\paragraph{Datasets}
We conduct experiments on four challenging autonomous driving datasets, including the Waymo~\cite{sun2020scalability,yan2024street}, KITTI~\cite{geiger2012we}, NOTR~\cite{yang2024emernerf} and Virtual KITTI 2 (VKITTI2)~\cite{cabon2020virtual}.
On Waymo, following~\cite{yan2024street}, we select eight challenging sequences with dynamic actors and complex background conditions and use the tracked boxes provided by~\cite{yan2024street} for experiments.
We select every fourth frame as the testing novel view and use the remaining frames as the training views.
We set the image resolution to $1066{\times}1600$ and evaluate both the novel view synthesis setting with the testing views and the image reconstruction setting with the training views.
On KITTI, following~\cite{turki2023suds,fischer2024multi}, we select three sequences and use the official tracklets for evaluation in the novel view synthesis setting.
We select every second frame as the testing novel view while the others are used for training.
The image resolution is set to $375{\times}1242$.
On NOTR, following~\cite{yang2024emernerf}, we employ two splits, namely, the static-32 and dynamic-32 splits, which consist of 64 multi-camera driving scene sequences with various challenges, \eg, cross-camera background variations, various dynamic actors, \etc.
The image resolution is set to $640{\times}960$.
On VKITTI2, following~\cite{zhou2024hugs}, we select two sequences for experiments and use the $50\%$ dropout rate evaluation protocol.
The image resolution is set to $375{\times}1242$.

\paragraph{Evaluation Metrics}
Following~\cite{yan2024street,yang2024emernerf,huang2024textit}, we use PSNR, SSIM~\cite{wang2004image} and LPIPS~\cite{zhang2018unreasonable} as the evaluation metrics.

\paragraph{Compared Methods.}
We compare our ArmGS approach with several state-of-the-art methods which can be categorized into two types:
(1) 3DGS-based methods, including StreetGS~\cite{yan2024street}, OmniRe~\cite{chen2025omnire}, SplatFlow~\cite{sun2025splatflow}, S$^3$Gaussian~\cite{huang2024textit}, 3DGS~\cite{kerbl20233d} and HUGS~\cite{zhou2024hugs};
(2) NeRF-based methods, including EmerNeRF~\cite{yang2024emernerf}, ML-NSG~\cite{fischer2024multi}, SUDS~\cite{turki2023suds}, NSG~\cite{ost2021neural}, StreetSurf~\cite{guo2023streetsurf} and MARS~\cite{wu2023mars}.

\subsection{Implementation Details}
\label{sec:sup_implement}
We implement our approach with Python and PyTorch.
We initialize driving scene Gaussians with LiDAR point clouds and SfM points extracted by COLMAP, and train our model with 30000 iterations using the Adam optimizer.
We adopt the training and evaluation splits for each dataset as described in Sec.~\ref{sec:exp-dataset}.
We set the initial position learning rate to $1.6e^{-4}$ and decay it to $1.6e^{-6}$.
The learning rates for rotation, scaling, opacity and spherical harmonics are set to $1e^{-3}$, $5e^{-3}$, $5e^{-2}$ and $2.5e^{-3}$, respectively.
We construct the local affine transformation learner with three linear layers, the global affine transformation learner with four linear layers and the class encoder with two linear layers, where ReLU activations are used between layers.
For the spatial-temporal encoder and the deformation head, we employ two linear layers with ReLU activations between layers.
By default, we use $\mathcal{F}^f$ for dynamic actor position encoding.
For training loss, we empirically set $\lambda_1{=}0.2$, $\lambda_2{=}0.01$, $\lambda_3{=}0.05$ and $\lambda_4{=}0.1$.
Gaussian splitting and merging for adaptive density control are performed every 100 iterations from 500 iterations and until 15000 iterations.

\begin{figure*}[t]
    \centering
    \includegraphics[width=0.9\textwidth]{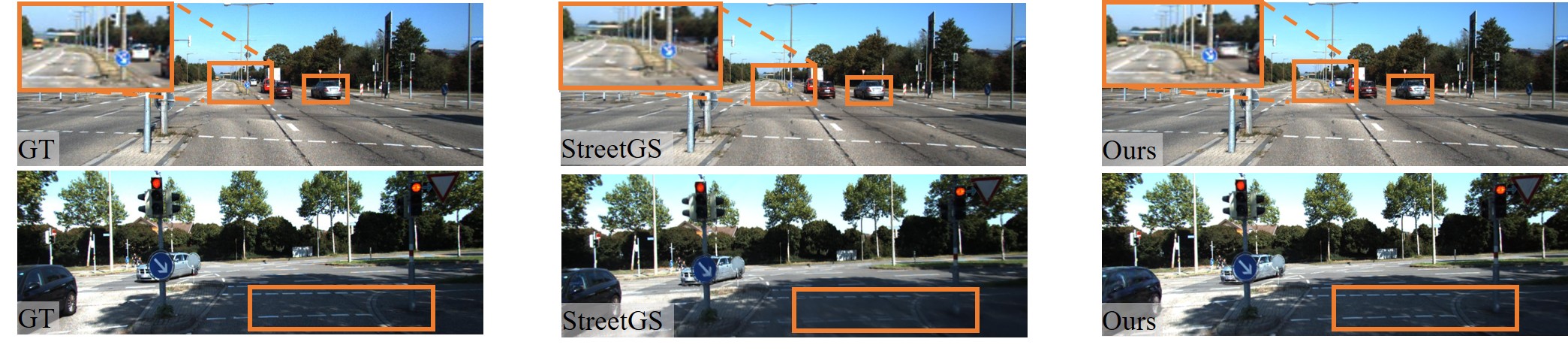}
    \caption{Qualitative comparison with the state-of-the-art methods on KITTI.
    We highlight some fine-grained details, \eg, vehicles and lane lines.
    }
    \label{fig:exp_vis_kitti}
\end{figure*}

\subsection{Evaluation on Waymo}
\label{sec:exp-waymo}
We present the quantitative results on Waymo in Tab.~\ref{tab:exp-sota-waymo-kitti}.
We can see that our approach achieves significantly better performance compared with the state-of-the-arts in both the image reconstruction and the novel view synthesis settings.
Specifically, for image reconstruction, our approach achieves the best PSNR of 38.1 dB, SSIM of 0.957 and LPIPS of 0.064, outperforming the state-of-the-art competitors, such as StreetGS, OmniRe and EmerNeRF.
For novel view synthesis, our approach still achieves the best results compared with the state-of-the-art methods. 
For instance, our approach achieves PSNR of 35.7 dB, which significantly surpass the second-best StreetGS by 1.1 dB.
Furthermore, in Fig.~\ref{fig:exp_vis_waymo}, we present some qualitative results of scene modeling under foggy, sunny, cloudy and rainy conditions on Waymo.
It can be seen that our approach renders more fine-grained details of the background scenes and dynamic actors.
For example, as shown in the first row, most methods fail to render the appearance change of traffic lights, while our approach render this fine-grained variation;
and as shown in the second row, our approach is able to model appearance of the distant vehicles, while other methods render results with artifacts.
Moreover, our experimental results show that the rendering speeds of the compared NeRF-based methods, \eg, EmerNeRF, are usually less than 1 FPS; in comparison, the 3DGS-based methods (including our approach) are able to achieve real-time rendering ($>$30FPS).

\subsection{Evaluation on KITTI}
\label{sec:exp-kitti}
We report the quantitative and qualitative results on the KITTI dataset in Tab.~\ref{tab:exp-sota-waymo-kitti} and Fig.~\ref{fig:exp_vis_kitti}, respectively.
Overall, our approach achieves significantly better performance compared with the state-of-the-art methods.
Specifically, from Tab.~\ref{tab:exp-sota-waymo-kitti}, we can see that our approach achieves the best PSNR of 30.3 dB, SSIM of 0.931 and LPIPS of 0.036, significantly outperforming the state-of-the-art methods.
Furthermore, from Fig.~\ref{fig:exp_vis_kitti}, we can see that compared with StreetGS, our approach is able to render better image quality with more fine-grained details, such as vehicles highlighted in the first row and the lane line highlighted in the second row.

\begin{figure}[t]
    \centering
    \includegraphics[width=0.49\textwidth]{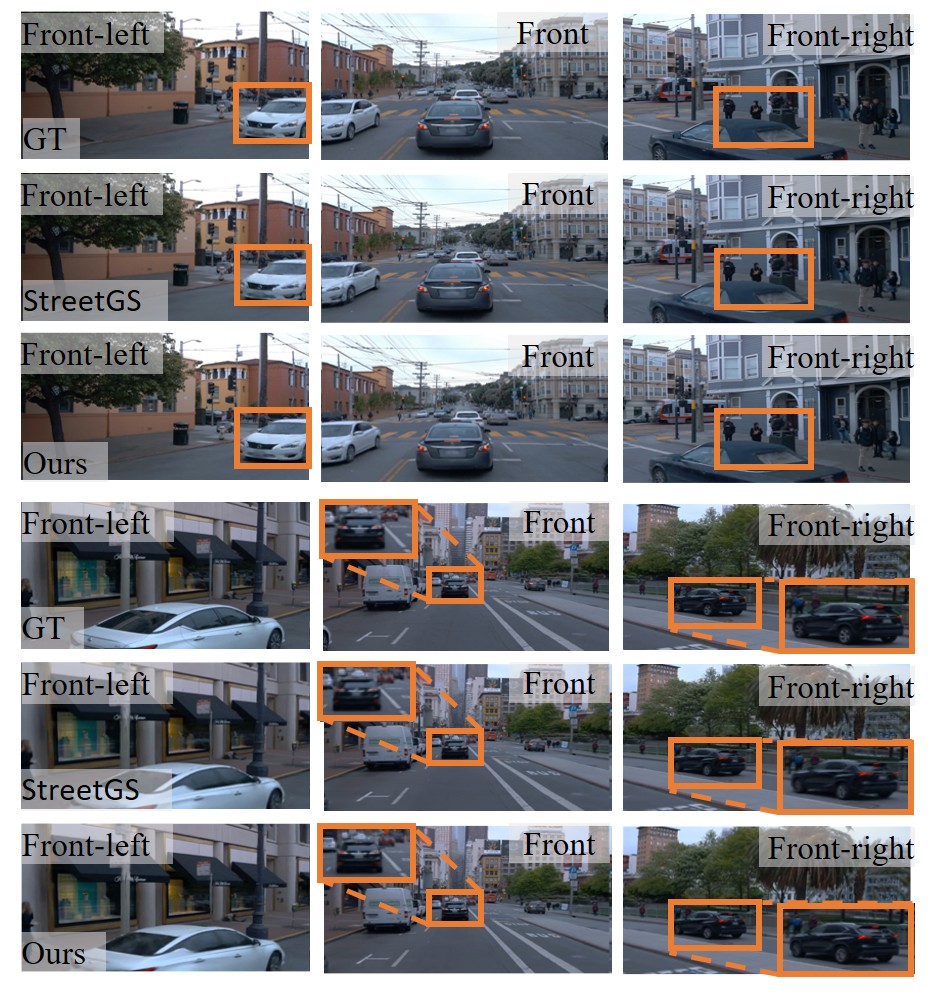}
    \caption{Qualitative comparison on NOTR.
    We highlight some fine-grained details, \eg, vehicles and pedestrians.
    }
    \label{fig:exp_notr}
\end{figure}

\begin{table}
    \small
    \centering
    \caption{Results on NOTR. PSNRs are reported.}
    \begin{tabular}{l|rr}
        \hline
        Method                  &  Static-32    &   Dynamic-32 \\
        \hline
        \hline
        StreetSurf~\cite{guo2023streetsurf}   & 26.2    & -   \\      
        EmerNeRF~\cite{yang2024emernerf}       & 29.1    & 28.9 \\  
        StreetGS~\cite{yan2024street}   & 29.6    & 29.7\\
        \hline
        ArmGS (Ours)                        & \textbf{30.5} & \textbf{30.5} \\
        \hline
    \end{tabular}
    
    \label{tab:exp-notr}
\end{table}

\subsection{Evaluation on NOTR}
\label{sec:exp-notr}
To further probe the upper bound reconstruction capability of our approach, we follow~\cite{yang2024emernerf} to conduct the scene reconstruction experiment on the NOTR dataset~\cite{yang2024emernerf}.
We report the quantitative and qualitative results in Tab.~\ref{tab:exp-notr} and Fig.~\ref{fig:exp_notr}, respectively.
As shown in Tab.~\ref{tab:exp-notr}, compared with the state-of-the-art StreetSurf~\cite{guo2023streetsurf}, EmerNeRF~\cite{yang2024emernerf} and StreetGS~\cite{yan2024street}, our approach achieves better performance on both the static-32 and dynamic-32 splits.
From Fig.~\ref{fig:exp_notr}, we can see that our approach is able to model fine-grained details of driving scenes across cameras; in comparison, StreetGS renders driving scenes with more artifacts, such as blurred vehicles.
These results show that our approach can faithfully model complex driving scenes and render high-quality results.

\begin{figure*}[t!]
    \centering
    \includegraphics[width=0.99\textwidth]{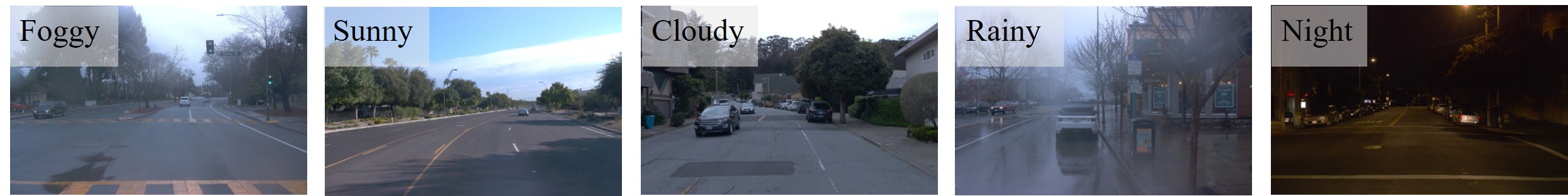}
    \caption{Urban scene modeling under diverse conditions using our approach.
    }
    \label{fig:exp_weather}
\end{figure*}

\subsection{Evaluation on VKITTI2}
\label{sec:exp-vkitti2}
Our approach is not limited to modeling real-world driving scenes, but can also be used to model synthetic driving scenes.
To show the performance of our approach for synthetic driving scenes modeling, we follow~\cite{zhou2024hugs} conducting experiments on the VKITTI2 dataset~\cite{cabon2020virtual}.
We report the results in Tab.~\ref{tab:exp-vkitti2}.
Although synthetic scenes usually have fewer appearance changes of scenes and objects across frames, our approach still achieves better results compared with the state-of-the-art HUGS.

\begin{table}
    \small
    \centering
    \caption{Results on VKITTI2.}
    \begin{tabular}{l|rrr}
        \hline
        Method                  &  PSNR$\uparrow$ & SSIM$\uparrow$ & LPIPS$\downarrow$ \\
        \hline
        \hline
        NSG~\cite{ost2021neural}               & 21.0    & 0.704   & 0.316 \\
        MARS~\cite{wu2023mars}                & 22.2    & 0.869   & 0.131   \\      
        3DGS~\cite{kerbl20233d}                & 20.6    & 0.892   & 0.103  \\
        HUGS~\cite{zhou2024hugs}               & 26.4    & 0.916     & 0.035 \\  
        \hline
        ArmGS (Ours)                                        & \textbf{31.1} & \textbf{0.957} & \textbf{0.016} \\
        \hline
    \end{tabular}
   
    \label{tab:exp-vkitti2}
\end{table}

\subsection{Ablation Study}
\label{sec:exp-ablation}

\paragraph{Dynamic Actor Modeling Effectiveness Analysis}
To evaluate the effectiveness of our dynamic actor level refinement, we follow \cite{yan2024street} projecting the 3D bounding box onto the 2D image plane to obtain the mask regions of moving objects and compute PSNR within the mask regions.
This mitigates the effect of background on dynamic actors when evaluating dynamic actors. 
As shown in Tab.~\ref{tab:exp-dynamic}, our approach achieves better results for the rendered dynamic actor regions compared with StreetGS and EmerNeRF.
When using class-wise actor encoding, our approach achieves similar performance.
When the dynamic actor level refinement is removed, the performance of our approach degrades in both reconstruction and novel view synthesis settings.
When replacing our design with the 4DGS HexPlane~\cite{wu20244d}, the result becomes worse.
We conjecture that, compared with our design, the 4DGS HexPlane requires structured factorization for spatial-temporal modeling which constrains its capability to capture variations of dynamic actors.

\begin{table}
    \small
    \centering
    \caption{Effectiveness analysis of dynamic actor modeling on Waymo.
    PSNRs of the moving object regions are reported.}
    \begin{tabular}{l|rr}
        \hline
        Method                  & Novel & Recon.        \\
        \hline
        \hline
        Ours                    & \textbf{30.9} & \textbf{35.8} \\
        Ours w/ class-wise actor encoding  & \textbf{30.9}     & 35.7 \\
        Ours w/o dynamic actor refinement & 30.7 & 35.4  \\  
        Ours w/ 4DGS HexPlane & 30.6 & 35.7 \\
        \hline         
        EmerNeRF & 21.7 & 26.8          \\
        StreetGS & 30.2 & 34.8          \\
        \hline
    \end{tabular}
    
    \label{tab:exp-dynamic}
\end{table}

\paragraph{Local and Global Appearance Refinement Effectiveness Analysis}
In Tab.~\ref{tab:exp-component}, we evaluate the effectiveness of the local composite Gaussian level refinement and the global image level refinement.
It can be seen that our approach with both local and global appearance refinement performs the best, while without using local level refinement or global level refinement, the performance of our approach degrades.
Besides, using HUGS appearance modeling in lieu of our design results in worse performance.

\begin{table}
    \small
    \centering
    \caption{Effectiveness analysis of local and global appearance refinement on Waymo.
    Results are in terms of PSNR.}
    \begin{tabular}{l|rr}
        \hline
        Method                           & Novel & Recon.        \\
        \hline
        \hline
        Ours                             & \textbf{35.7} & \textbf{38.1} \\
        \hline
        Ours w/o local level refinement  & 35.0 & 36.8    \\
        Ours w/o global level refinement & 35.4 & 37.7  \\
        Ours w/ HUGS Appearance          & 34.8  & 36.5 \\
        \hline
    \end{tabular}
    
    \label{tab:exp-component}
\end{table}

\begin{table}[t!]
  \small
  \centering
  \caption{Ablation analysis of more components.}
  \begin{tabular}{l|c}
    \hline
    Method                    & PSNR$\uparrow$       \\
    \hline
    ArmGS                     & 38.1        \\
    \hline
    without LiDAR depth loss    & 37.9       \\
    without LiDAR initial points        & 37.5       \\
    without sky GS model and mask & 37.6           \\
    without actor pose optimization    &  37.7  \\
    add camera pose optimization    & 38.1        \\
    \hline    
  \end{tabular}
  
  \label{tab:exp-more-ablation}
\end{table}

\paragraph{Ablation Analysis of More Components}
In Tab.~\ref{tab:exp-more-ablation}, we report ablation analysis of more components.
We can see that:
(1) LiDAR depth loss has a minor impact on our approach;
(2) Without LiDAR initial points, the performance of our approach only slightly decreases;
(3) Sky Gaussians can slightly affect the model performance;
(4) Actor pose optimization helps the model to achieve better performance;
(5) Adding additional camera pose optimization to our approach does not bring performance improvement.

\paragraph{Urban Scene Modeling under Diverse Weather Conditions}
Real-world urban scenes often involve diverse weather conditions. 
Although our approach is not specifically designed for urban scene modeling under extreme weather conditions, our approach still performs well for different urban scene conditions, \eg, foggy, sunny, cloudy, rainy and nighttime.
Fig.~\ref{fig:exp_weather} shows some results of our approach for urban scene modeling under diverse conditions.

\section{CONCLUSION}
\label{sec:conclusion}
This work presents a composite Gaussian appearance refinement approach to urban environments modeling.
The key idea is to learn a set of transformation parameters to refine 3D Gaussians from multi-level granularities, ranging from local Gaussian level to global image level and dynamic actor level.
Our thorough experiments on four autonomous driving datasets demonstrate that our approach is capable of rendering more fine-grained details of driving scenes across frames and camera viewpoints, achieving superior performance compared with the state-of-the-art methods.

\bibliographystyle{IEEEtran}  
\bibliography{ref}

\end{document}